# Data Pre-Processing and Evaluating the Performance of Several Data Mining Methods for Predicting Irrigation Water Requirement

**Mahmood A. Khan**
School of Environmental Sciences, Charles Sturt University
Wagga Wagga 2678, NSW, Australia
makhan23.dr@gmail.com

**Md Zahidul Islam**
School of Computing and Mathematics, Charles Sturt University
Bathurst 2795, NSW, Australia
zislam@csu.edu.au

**Mohsin Hafeez**
International Water Management Institute (IWMI)
M.hafeez@cgiar.org

*Recent drought and population growth are planting unprecedented demand for the use of available limited water resources. Irrigated agriculture is one of the major consumers of fresh water. Large amount of water in irrigated agriculture is wasted due to poor water management practices. To improve water management in irrigated areas, models for estimation of future water requirements are needed. Developing a model for forecasting irrigation water demand can improve water management practices and maximise water productivity. Data mining can be used effectively to build such models.*

*In this study, we prepare a dataset containing information on suitable attributes for forecasting irrigation water demand. The data is obtained from three different sources namely meteorological data, remote sensing images and water delivery statements. In order to make the prepared dataset useful for demand forecasting and pattern extraction we pre-process the dataset using a novel approach based on a combination of irrigation and data mining knowledge. We then apply and compare the effectiveness of different data mining methods namely decision tree (DT), artificial neural networks (ANNs), systematically developed forest (SysFor) for multiple trees, support vector machine (SVM), logistic regression, and the traditional Evapotranspiration (ETc) methods and evaluate the performance of these models to predict irrigation water demand. Our experimental results indicate the usefulness of data pre-processing and effectiveness of different classifiers. Among the six methods we used, SysFor produces the best prediction with 97.5% accuracy followed by decision tree with 96% and ANN with 95% respectively by closely matching the predictions with actual water usage. Therefore, we recommend using SysFor and DT models for irrigation water demand forecasting.*

## 1. INTRODUCTION

Water availability plays an important role in irrigated agricultural. Water scarcity is rapidly becoming a major issue for many developed and developing countries of the world, which is a serious threat and leads to emergence of food crisis (IWMI, 2009). As the scarcity of the water increases, the demand for managing available water resources becomes crucial. In particular, a recent drought in Australia has made prominent the need to manage agriculture water more wisely. It is reported that, more than 70% of available water in Australia and 70% to 80% of water Worldwide is currently being used by irrigated agriculture (Khan *et al*, 2009; Khan *et al*, 2011; IWMI, 2009). Due to recent drought, climate change, population growth and increasing demand for domestic and industrial water requirement, preserving sufficient amount of freshwater for agricultural production will become increasingly difficult. Since all the existing water resources are fully utilised and drawing of more water is impracticable, therefore the best alternative is to increase the water productivity (Khan *et al*. 2011). Studies report that, water delivered for irrigation is not always efficiently used for crop production, on an average 25% of water is wasted due to inefficient water management practices (FAO, 1994; Smith, 2000).

To improve water management practices and maximise water productivity, application of various hydrological and data driven models using data mining methods have become very essential (Khan *et al*. 2012). In the current situation, models to predict future water requirements based on data mining techniques can be useful. Ullah *et al*. (2011) suggests that, to developing a model for water demand forecast, it is essential to understand the behaviour of the irrigation system in the past, the current land use trends and the behaviour of future hydrological attributes such as (rainfall, evapotranspiration, seepage, etc.). Having an accurate and reliable Irrigation water demand forecasting model based on hydrological, meteorological and remote sensing data can provide important information to agriculture water users and managers (Pulido-Calvo *et al*, 2009; Zhou *et al*, 2002; Alvisi *et al*, 2007)

In recent years, according to Pulido-Calvo *et al* (2003) data mining techniques are increasingly being applied in the field of hydrology for developing models to predict various hydrological attributes such as rainfall, pan evapotranspiration, flood forecasting, weather forecasting etc. However, these techniques are not used to predict irrigation water demand. Data mining discovers new and practically meaningful information from large datasets. Unlike any typical statistical methods, data mining techniques explores interesting and useful information without having any pre set hypotheses. These techniques are more powerful, flexible and capable of performing investigative analysis (Olaiya *et al*, 2012). Zurada *et al*, (2005) says data mining uses a number of analytical tools such as decision trees, neural networks, fuzzy logic, rough sets, and genetic algorithms to perform classification, prediction, clustering, summarisation, and optimisation. The most common tasks among these are classification and prediction which we carried out in this study.

The aim of this study is to a) prepare and pre-process the dataset b) apply and compare the effectiveness and accuracies of different data mining models on pre-processed datasets in determining irrigation water requirement. Since, there is a strong need for data pre-processing to get good quality results, we pre-process our dataset using a novel approach called Reference Evapotranspiration Based Estimate, which is based on Reference Evapotranspiration ($ET_c$), a comprehensive explanation can be found in section 3.1 of this paper.

We build models on pre-processed datasets based on five data mining techniques namely decision trees, artificial neural networks, systematically developed forest (SysFor), support vector machine, logistic regression, and traditional $ET_c$ based method. Our experimental results indicate that as a result of data pre-processing the quality of training dataset increases significantly and a minor difference in the prediction accuracies between different data mining techniques. However, among the five different techniques/models the prediction performance of multiple decision tree technique Sysfor is found to be the best followed by Decision Tree and ANN. An implementation of SysFor is freely available in WEKA (https://www.cs.waikato.ac.nz/ml/weka/) and a demonstration video on how to use SysFor in WEKA is available at https://www.youtube.com/watch?v=DQKKdAahDgE&t.

This paper is organised as follows, section 2 describes the study area, followed by data pre-processing and classification methods used in this study are described in section 3. Experimental results are explained in the Section 4, Section 5 concludes the paper with some suggestions for future work.

## 2. STUDY AREA

In this study, Coleambally Irrigation Area (CIA) is selected as our study area. CIA is one of the most modernized irrigation areas in the Murray and Murrumbidgee river basins of Australia. CIA is situated approximately 650km south-west of Sydney in the Riverina District of New South Wales which falls under lower part of Murrumbidgee River Catchment as shown in Figure 1. CIA contains approximately 79,000ha of intensive irrigation area and 325,000ha of the Outfall District area, supplying water to 495 irrigation farms (CICL, 2011). Because of the recent drought in the last decade, there is a significant decline in the average water allocation to the farmers of CIA. Due to declining water allocation and changing weather patterns, CIA requires new management measures for using water efficiently and increase water productivity.

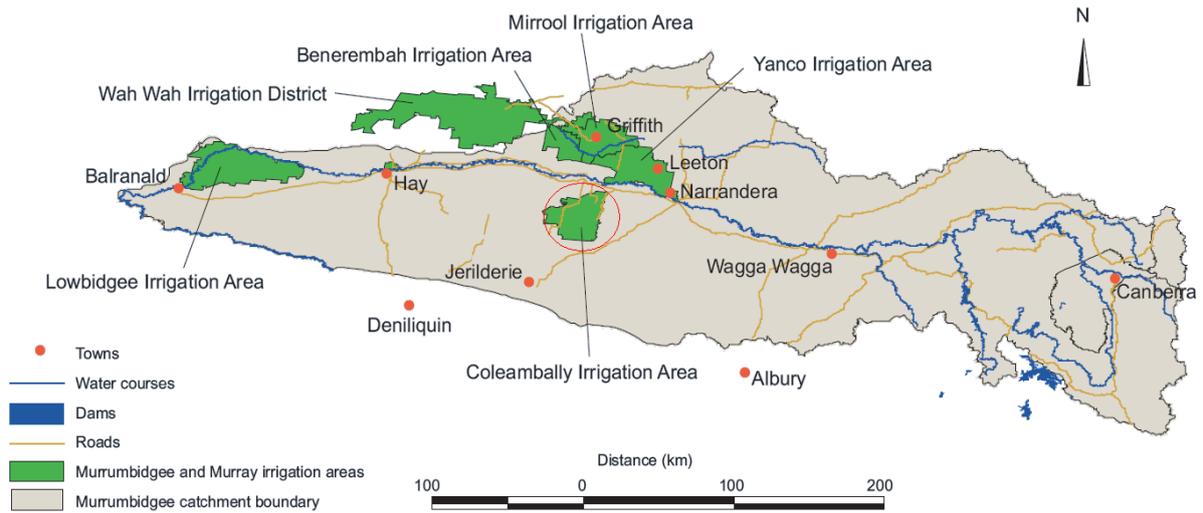

**Figure 1: Location of Coleambally Irrigation Area and Other Major Irrigation Areas in Murrumbidgee Catchment**

## 3. DATA AND DESCRIPTION OF METHODS
### 3.1 Data Collection and Data Pre-processing

To build the training dataset, we collect data from three different sources. The first source is the water delivery statements that are obtained from CICL and provides us with the information about total water usage for a crop growing season by each farm. The second source is the meteorological data that are obtained from the installed weather stations in the study area, and the third source is spatial data that are of two types a) Land Use and Land Cover images, which provide us with information about the crops grown and the cropping area b) Soil Type images that gives us information about the different soil types associated with the farms in the study area.

We choose those attributes that have significant influence on crop water usage and the data for which is available throughout the whole cropping seasons. For example, "cropping stage" has strong relationship with crop water usage. However, due to non-availability of reliable past data on cropping stage we do not add this attribute in our dataset. Similarly the attributes for which future data can be difficult to obtain are not chosen.

Our dataset contains historical data composed of attributes on various weather parameters such as Maximum and Minimum Temperature (T-Max & T-Min), Wind speed, Humidity, Rainfall, and Solar Radiation combined with Soil Type, Crop Type and Crop Water Usage (see Table 1). Attribute crop water usage in Table 1 is termed as "class" attribute and all others as "non-class" attributes.

| Non-Class Attributes | | | | | | | | Class Attribute |
|---|---|---|---|---|---|---|---|---|
| T-max (ºC) | T-min (ºC) | Humidity (%) | Wind Speed km/day | Rainfall (mm) | Solar Radiation (MJ/m^2) | Soil Type | Crop Type | Crop Water Usage (ML/ha/day) |
| 18.1 | 3.8 | 80 | 122 | 0.2 | 9.5 | SMC | Barley | 0.01-0.05 |
| 16.4 | 6.7 | 48 | 481 | 0.0 | 16.6 | RBE | Wheat | 0.06-0.10 |
| 30.1 | 14.0 | 65 | 275 | 0.0 | 24.7 | SMC | Rice | 0.11-0.15 |
| 30.7 | 15.9 | 58 | 257 | 0.0 | 29.3 | SMC | Corn | 0.06-0.10 |

**Table 1: Example of our training dataset, Crop Water Usage is the Class attribute.**

In this study, we consider the dataset as a two dimensional table where columns are attributes (categorical & numerical) and rows are records. Each record holds the daily average values of the corresponding attributes. Attributes such as soil type and crop type are classified as categorical whereas all other non-class attributes as numerical.

While preparing the dataset we also face a couple of challenges as follows. Our first data source is Water Delivery Statement, which only provides us with the information on the date and amount of water supplied/delivered to a farm. Note that a farm does not take water supply every day. Instead, it takes a delivery of water on a day and uses the water for a period of time. Therefore, from the water delivery statement we only get information on the amount of water delivered on any particular date. From the water delivery statement it is not possible to estimate the exact amount of crop water usage for a particular day. However, in order to obtain an accurate relationship between the non-class and the class attribute we need daily crop water usage for each record of the training dataset. We propose a data pre-processing approach called Reference Evapotranspiration based Pre-processing (REP) for estimating the daily crop water usage of a farm. Moreover, we compare the proposed REP technique with another possible approach called Equal Water Distribution based pre-processing (EWD), both approaches are explained as follows.

In Equal Water Distribution technique (EWD) we divide the volume of water delivered to a farm by the number of days between two consecutive deliveries. Therefore, we get an average water usage per day. However, if we divide the water usage evenly, then water usage remains the same for each day regardless of weather conditions. Since actual crop water usage has a strong relationship with the weather condition, we need more accurate water usage data in our dataset. Apply a classification model on precise dataset will explore accurate relationship between the non-class attributes and the class attribute (Crop Water Usage).

In Reference Evapotranspiration based Pre-processing (REP), we estimate daily water usage more accurately than EWD. Here, we take reference evapotranspiration (ETo) factor into consideration. Crop water usage can be calculated through Evapotranspiration (ET) which is the product of crop coefficient Kc and reference evapotranspiration (ETo) (Al Kaise et al, 2009). Each crop has a constant crop coefficient value for a specific growth stage.

We calculate the crop water usage of a particular day as follows. Let, n be the number of days between two consecutive water deliveries for a farm, WT be the amount of water delivered during the delivery at the beginning of the n days, and Wi be the water usage in the ith day. We obtain the daily ETo values, for all n days, from our Automatic Weather Stations (AWS) placed in different locations of our study area. We next calculate the coefficient xi for the ith day, where

$$x_i = (ET_o^i)/(\Sigma_{(j=1)}^n ET_o^j)$$

$ET_o^i$ is the $ET_o$ of the *i*-th day. Finally, $W_i$ is calculated by multiplying $x_i$ and *WT*, i.e. $W_i = x_i \times WT$.

This pre-processing approach is superior to the earlier approach. Unlike EWD approach here we do not use the average water usage for the days having different weather conditions. It estimates water usage as accurate as possible for each day and thereby uses each record of training dataset.

Our second data source is meteorological data i.e., historical data on weather parameters such as T-Max, T-Min, Wind speed, Humidity, Rainfall, and Solar Radiation. The third and the final data source is spatial data, which gives information about seasonal land use (cropping pattern) and soil types associated with the farms. Using the spatial maps pre-processed from satellite images crop type, cropping area and soil type information are extracted.

### 3.2 Classification Methods

All the methods/techniques used to predict water demand forecast in this study are well known and well established. Therefore, we explain only the basic functionalities of each method, without explaining the mathematical descriptions of the underlying algorithms. For more information relating to any specific algorithm on decision tree, artificial neural networks, support vector machine, systematically developed forest (SysFor) and logistic regression refer to (Quinlan, 1993; Islam, 2010; Khan et al, 2011; Cancelliere et al, 2002; Yang et al, 2006; Han & Kamber, 2001; Vapnik 1995; Islam and Giggins 2011; Christensen, 1997). A brief explanation of the methods is as follows.

#### 3.2.1 Decision Tree (DT)

Decision trees are a powerful tool for data classification. Decision tree learns from the training dataset and apply the learned knowledge on the testing dataset to find the hidden relationships between the classifying (class) and classifier (non class) attributes. A class attribute is an attribute of the dataset, which contains the values that are possible outcomes of the record. A decision tree analyses a set of records whose class values are known (Quinlan, 1996). In other words, a decision tree explores patterns also known as logic rules from any dataset (Islam, 2010). By using the rules generated by a decision tree the relationship between the attributes of a dataset can be extracted. Each rule represents a unique path from the root node to each leaf of the tree.

Decision trees are made of nodes and leaves as shown in Figure 2, where each node in the tree represents an attribute and each leaf represents the value for the records belonging to the leaf (Khan et al, 2011; Han and Kamber, 2001). The concept of information gain is used in deciding the best suitable attribute for a node. The functionality of the decision tree is based on C4.5 algorithm. C4.5 takes a divide and conquers approach to build a decision tree from a training dataset using the principle of information gain (Quinlan, 1993).

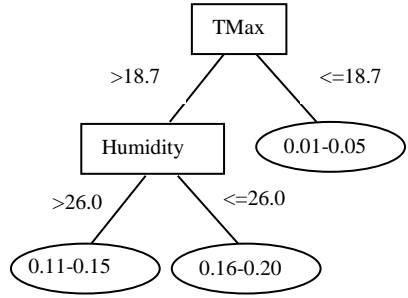

**Figure 2: An Example of a Decision tree generated from our dataset**

### 3.2.2 Artificial Neural Networks (ANN)

Artificial Neural Network (ANN) is a data processing and classification model that is inspired by the biological neural network. ANN learns the non-linear relationships, trends and patterns from training dataset and uses the knowledge for predicting the class values of unseen datasets (Cancelliere et al, 2002; Yang et al, 2006).

Interconnection strengths known as weights are used to store the gained knowledge. Weights of the neurons in ANN are computed during the training process. Based on the nature of the datasets an appropriate network can be selected, where a user/data miner can choose number of layers and number of nodes in each layer of the network. In hydrological modelling most ANNs are trained with single hidden layer (Dawson and Wilby, 2001; de Vos and Rientjes, 2005) as reported by Wu et al, (2010). The ANN model is based on error minimisation principle. Training of the model can be carried out in two ways; supervised and unsupervised learning (Han and Kamber 2001).

One of the most popular and commonly used ANN architectures is multilayer feed-forward neural network as shown in Figure 3, which is also called as multilayer perceptron (Muttil and Chau, 2006). In a multilayer perceptron network there is an input layer, an output layer and one or more hidden layers. These layers extract patterns from a dataset and use the learned patterns to predict class values of new records. The nodes in the input layer pass the processed information to the computational nodes in a forward direction Wang et al, (2009). The hidden layer is also responsible for resolving the nonlinearity between the input and output attributes of the dataset (Ambrozic and Turk, 2003; Cancelliere et al, 2002; Safer, 2003).

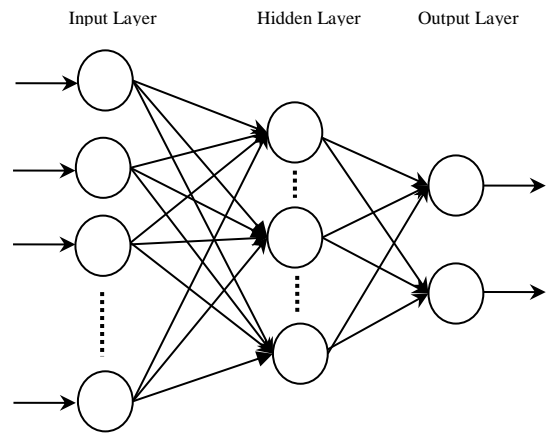

**Figure 3: Architecture of three - tier feed forward neural network.**

### 3.2.3 Systematically Developed Forest of Multiple Trees (SysFor)

SysFor is a multiple tree building technique based on the concept of gain ratio. This technique is developed by Islam and Giggins, (2011). The purpose of building multiple trees is to gain better knowledge through the extraction of multiple patterns. We explain this technique in a step by step fashion.

In the first step, a set of good attributes and their split points are identified based on user defined goodness (gain ratio) and separation values. Islam and Giggins (2011) says, a numerical attribute can be chosen more than once within the set of good attributes, if it has higher gain ratios with different split points that are not close to each other. After the set of good attributes are selected and if the size of the good attributes is less than a user defined number of tree, then in the next step (step 2) SysFor builds the tree using each good attribute as the root attribute of the tree, and build as many trees as number of good attributes. Else it builds user defined number of trees from the set of good attributes as the root attribute.

If the number of trees build in this step are still less than the user defined number of trees, then SysFor in the next step (step 3) build more trees until user defined number is met by using alternative good attributes at the next level of the tree i.e. at level 1 of the tree generated in the previous step (step2). In this step (step 3) the algorithm first uses the root attribute of the first tree built in step 2 in order to split dataset into horizontal partition. The algorithm, then selects a new set of good attributes, their respective split points and a set of gain ratios for each horizontal partition. Based on these set of good attributes the algorithm builds a tree from each partition and the trees are joined by connecting their roots (at level 1) to the root (at level 0) of first tree build in step 2. This process of building more trees continues until user defined number of trees are generated/build. Example trees generated in SysFor are shown in Figure 4a, 4b.

After Systematic forest of multiple trees is generated as to predict the class values of unseen records we follow voting system proposed by Islam and Giggins, (2011) called SysFor Voting-2. In this voting system, we find all the leaves from all the trees the record falls into. Then the leaf with highest accuracy is determined (based on maximum number records with same class values to total number of records) and finally the majority class value of the leaf is chosen as the predicted class value of the record.

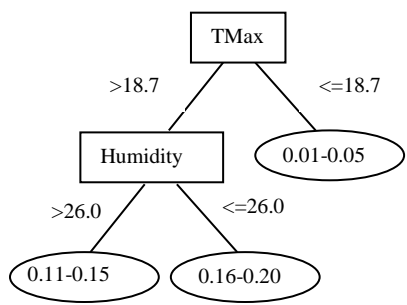
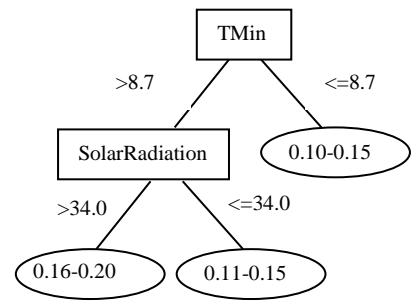

**Figure 4a: Tree generated in SysFor based on first good attribute**

**Figure 4b: Tree generated in SysFor based on second good attribute**

### 3.2.4 Support Vector Machine (SVM)
Support vector machine is a state of the art neural network methodology based on statistical learning (Vapnik, 1995, Wang et al, 2009). An SVM is an algorithm for maximizing a particular mathematical function with respect to a given dataset. The basic concepts behind the SVM algorithm are i) the separating hyperplane, ii) the maximum-margin hyperplane, iii) the soft margin and iv) the kernel function. A support vector machine constructs a hyperplane or set of hyperplanes in a high- or infinite-dimensional space, which can be used for classification. In general, a good separation is achieved by the hyperplane that has the largest distance to the nearest training data point of any class as shown in Figure 5 which exhibits the basic concept of support vector machine. From Figure 5 is it can be seen that the optimal hyperplane separates the positive and negative points from the dataset with a maximum margin, indicating the maximum distance to hyperplane from closest positive and negative data points.

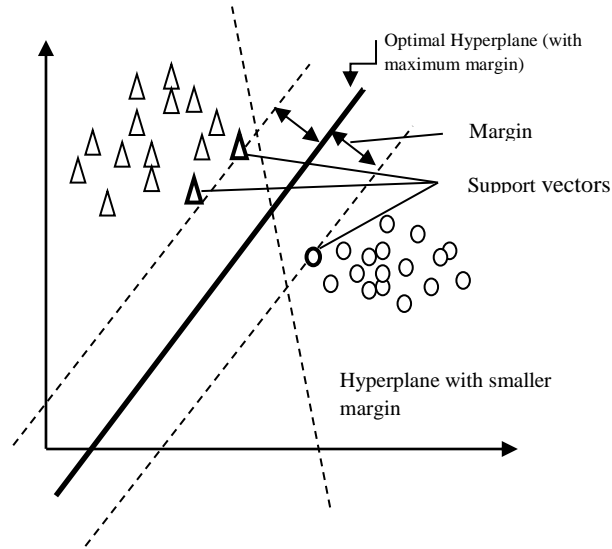

**Figure 5: Basic concept of support vector machine**

### 3.2.5 Logistic Regression
The main goal of logistic regression model is to predict the label t of a new given data point x based on the learning from the training dataset. Logistic regression can be of two types 1) Simple Logistic Regression and ii) Multiple Logistic regression. Simple logistic regression is used to predict the class value, given it is categorical and has only two possible outcomes such as (male/female), whereas the multiple logistic regressions can be used to predict the class value consisting of three or more possible outcomes. Logistic regression is a capable probabilistic binary classifier (Christensen, 1997). A logistic regression model helps us assess probability from which the outcomes will be chosen.

It is evident from the literature that the logistic regression is used extensively in numerous disciplines such as, in the field of medical and social sciences, marketing applications etc (Pearce& Ferrier, 2000). (Zurada, 2005), states that logistic regression models are designed to predict one class value at a time and they are assumed as simplest feed forward neural networks containing only two layers input and output.

### 3.2.6 Evapotranspiration ($ET_c$) based Prediction

$ET_c$ can be broadly defined as crop water usage. Crop Evapotranspiration ETc is calculated using crop coefficient $K_c$ (for a crop type and cropping stage) and reference evapotranspiration ($ET_o$). The empirical formula to calculate $ET_c$ is $ET_c = K_c \times ET_o$ (FAO 56), and this is commonly used globally to estimate water demand. The crop coefficient method was developed for the agriculture users to calculate $ET_c$ which helps them in making irrigation management decisions.

## 4. EXPERIMENTAL RESULTS

In order to evaluate the performance of our data pre-processing techniques we build two training datasets $D_1$ and $D_2$. The dataset $D_1$ is based on our Equal Water Distribution (EWD) approach and $D_2$ is based on our Reference Evapotranspiration based Pre-processing (REP).

### 4.1 Application of Classification methods on dataset $D_1$

A decision tree classification method on dataset $D_1$ is applied. We divide $D_1$ into two parts training and testing. The decision tree is built on training dataset to extract the relationship between the non-class and class attributes and applied on testing dataset to check the prediction accuracy of unseen records. We implement C4.5 algorithm to generate a decision tree.

Similarly, an ANN is built on $D_1$ using the three tier feed-forward architecture with back propagation. To build an ANN, we divide the datasets into three parts; 70%, 20% and 10% for training, validating and testing, respectively. Training of the network is performed using two different network topologies, firstly by using 1 hidden layer having 8 nodes, and secondly by using 1 hidden layer having 6 nodes. Both the networks are trained for 30000, 50000 and 70000 learning iterations. The network produced by 1 hidden layer with 8 nodes for 30000 learning iterations produces better results. The ANN is built using EasyNN plus V14.0 software (available from http://www.easynn.com/).

We also build SysFor on our dataset $D_1$, by considering user defined number of trees to be 5 and follow SysFor voting 2 for predicting the unseen records.

Finally we train and test SVM and Logistic regression using WEKA 3.6.2 which is available at http://www.cs.waikato.ac.nz/~ml/weka/ and very popularly used tool for performing different data mining tasks.

The performance evaluation of the models on dataset $D_1$ is carried out by comparing the prediction accuracies. The prediction accuracy check is performed using a 3 fold cross validation method. This is a method of testing the accuracy by dividing the dataset in three equal parts also called as folds, where two parts of the dataset are used for training and the third part is used for testing. This process is continued 3 times so that each part of the dataset is used once for testing. Dataset D1 has 6070 records in total where 2023 records are used for testing in each cross validation. Table 2 displays the prediction accuracies of each fold for all the above mentioned models used in our experiment on dataset $D_1$.

| Folds | Classification Models | | | | |
|---|---|---|---|---|---|
| | DT | ANN | SysFor | SVM | Logistic Regression |
| 1 | 34.6 | 34.7 | 44.2 | 33.1 | 31.6 |
| 2 | 46.6 | 41.0 | 53.6 | 39.7 | 33.8 |
| 3 | 51.3 | 39.1 | 55.3 | 34.6 | 38.2 |
| Average (%) | 44.1 | 38.2 | 51.03 | 35.8 | 34.5 |

Table 2: Prediction accuracies of different models based on 3 folds cross validation on dataset $D_1$

Table 2 indicates that multiple decision tree technique SysFor has performed the best among all other techniques with an accuracy average of 51%, followed by decision tree 44.1%, ANN 38.2%, SVM with an average of 35.8% and the least performed logistic regression with 34.5%.

**4.2 Application of Classification methods on dataset $D_2$**

We now apply the same classification methods to our dataset $D_2$. Dataset $D_2$ has 1500 records where 500 records are used for testing in each cross validations Table 3 displays the performance of different classification methods on dataset $D_2$.

It is evident from Table 3 that the performance of multiple decision tree technique Sysfor again is better among all the other techniques, followed by decision tree and SVM. SysFor records 78% prediction accuracy while DT and SVM exhibit an accuracy of 74% and 64% respectively. The accuracy of ANN and logistic regression were recorded low. It can be noted that ANN has performed better than SVM on dataset $D_1$. However it did not repeat the same with dataset $D_2$.

| Folds | Classification Model | | | | |
|---|---|---|---|---|---|
| | DT | ANN | SysFor | SVM | Logistic Regression |
| 1 | 72.6 | 59.3 | 75.5 | 63.2 | 57.1 |
| 2 | 74.5 | 60.7 | 83 | 62.1 | 53.7 |
| 3 | 73.8 | 62 | 77.9 | 67.1 | 56.7 |
| Average (%) | 74 | 61 | 78 | 64 | 56 |

Table 3: Prediction accuracies of different models based on 3 folds cross validation

The results displayed in Table 2 and 3 clearly indicate the effectiveness of our data pre-processing technique REP, based on the knowledge in irrigation engineering and data mining. Moreover, the result also indicates the appropriateness of the attributes selected using three different sources namely water delivery statement, meteorological data and remote sensing processed images obtained from satellite data. We also compare the accuracies of the experimented models with the accuracy of traditional approach of estimating water requirement, which is based on actual crop evapotranspiration (ETc).

In addition to accuracy test we also compare the closeness of actual water consumed by the crop to the water predicted by the above mentioned models for summer season of the year 2008/09. Table 4 shows a comparison between the actual water usage, water usage predicted by the decision tree, ANN, SysFor, SVM, Logistic regression and traditional ETc based approach for all the 22 nodes of CIA.

All 6 models are applied on every farm of CIA to obtain the water demand for a whole cropping season. The water demand for each node is calculated by adding the water demand predicted for the farms belonging to the node. The accuracy of closeness for actual and predicted water is calculated as follows

$$\text{Accuracy} = 1 - \left(\frac{|\text{Actual-Predicted Water Usage}|}{\text{Actual Water Usage}}\right) \times 100\%$$

From Table 4 it is clear that the water demand predicted by SysFor is more closely matching the actual water consumed. The accuracy of closeness is found to be 97.5% which suggest a higher closeness of prediction made by the model. The accuracy of SysFor is followed by decision tree and ANN whose closeness were found to be 96% and 95% which are also considered to be very high. However, in few nodes such as Yamma1 and Boona 2 the prediction of SysFor was worse than decision tree and ANN. In majority of the nodes the performance of SVM, Logistic regression and ETc was behind the performance of Sysfor, decision tree and ANN.

Moreover, in few nodes such as "Coly 7", "Bundure_Main" and "Bundure 7_8", the actual water usage is significantly lower than the water usage predicated by all the models. This is because only a few farms of the nodes were irrigating during the season. The farms stopped irrigation for some reason half way through the season as it is evident from the water delivery statement. The node "Coly 10" does not have any irrigation for the cropping season. We exclude results of these nodes while calculating the accuracy of the models. In Table 4 the rows representing the above said nodes are shaded to highlight the exclusion of these nodes.

Figure 6 and Figure 7 displays the basic comparison between actual and predicted water usage. Figure 6 show the positive (predicted more) and negative (predicted less) predictions to actual water usage for all 22 nodes of CIA from all six models. It is evident from Figure 6 that the bars representing SysFor and DT are shorter for all nodes compared to the longer bars representing other models. Therefore, we can say that the predictions made by SysFor and DT are close to actual water usage. Similarly, the scatter plots in Figure 7 shows the actual and predicted water usage made by all the models experimented in this study.

| Node | Predicted Water Usage | | | | | | |
|---|---|---|---|---|---|---|---|
| | Actual Water Usage (ML) | Decision Tree (ML) | ANN (ML) | SysFor (ML) | SVM (ML) | Regression (ML) | $ET_c$ (ML) |
| Coly 1_2 | 407 | 344 | 316 | 379 | 417 | 428 | 284 |
| Coly 3 | 1292 | 1203 | 1210 | 1155 | 1278 | 1417 | 777 |
| Coly 4 | 800 | 746 | 1262 | 759 | 841 | 931 | 570 |
| Coly 5 | 879 | 945 | 1383 | 1001 | 1110 | 1228 | 666 |
| Coly 6 | 4359 | 4158 | 3807 | 4464 | 4891 | 5266 | 3235 |
| Coly 7 | 82 | 220.5 | 245 | 231 | 256 | 283 | 157 |
| Coly 8 | 785 | 802 | 830 | 850 | 1084 | 1139 | 875 |
| Coly 9 | 4501 | 4297 | 4394 | 4317 | 4801 | 5211 | 3232 |
| Coly 10 | 0 | 0 | 0 | 0 | 0 | 0 | 0 |
| Coly 11 | 2262 | 2877.5 | 3104 | 2581 | 2996 | 3139 | 2264 |
| Tubbo | 696 | 630 | 814 | 645.7 | 716 | 792 | 444 |
| Boona 1 | 1201 | 1069 | 1692 | 1189 | 1323 | 1424 | 791 |
| Boona 2 | 418 | 429 | 531 | 550 | 720 | 797 | 259 |
| Boona 3 | 2438 | 2101 | 2268 | 2341 | 2585 | 2713 | 1652 |
| Yamma Main | 4299 | 3732 | 4542 | 4375 | 4921 | 4966 | 3098 |
| Yamma 1 | 3333 | 3364 | 3100 | 3940 | 5558 | 5558 | 3085 |
| Yamma 2_3_4 | 2926 | 3045 | 3207 | 3180 | 4479 | 4370 | 2772 |
| Bundure Main | 87 | 493 | 650 | 646 | 726 | 745 | 419 |
| Bundure 3 | 763 | 768 | 636 | 798 | 897 | 901 | 653 |
| Bundure 4 | 1597 | 1384 | 1421 | 1387 | 1560 | 1532 | 897 |
| Bundure 5_6 | 961 | 798 | 660 | 836 | 935 | 941 | 677 |
| Bundure 7_8 | 133 | 378 | 504 | 396 | 440 | 486 | 268.5 |
| **Coleambally Irrigation Area** | **33917** | 32692.5 | 35177 | **34747.7** | 41112 | 42753 | 26231 |

Table 4: Comparison of water usage predicted by different models to actual water usage for all nodes of CIA

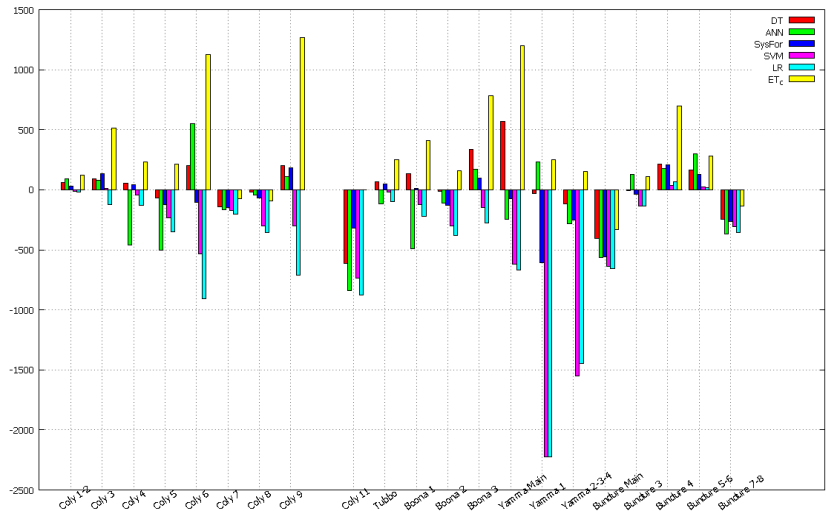

**Figure 6: Positive and Negative difference between actual and predicted water usage made by different models**

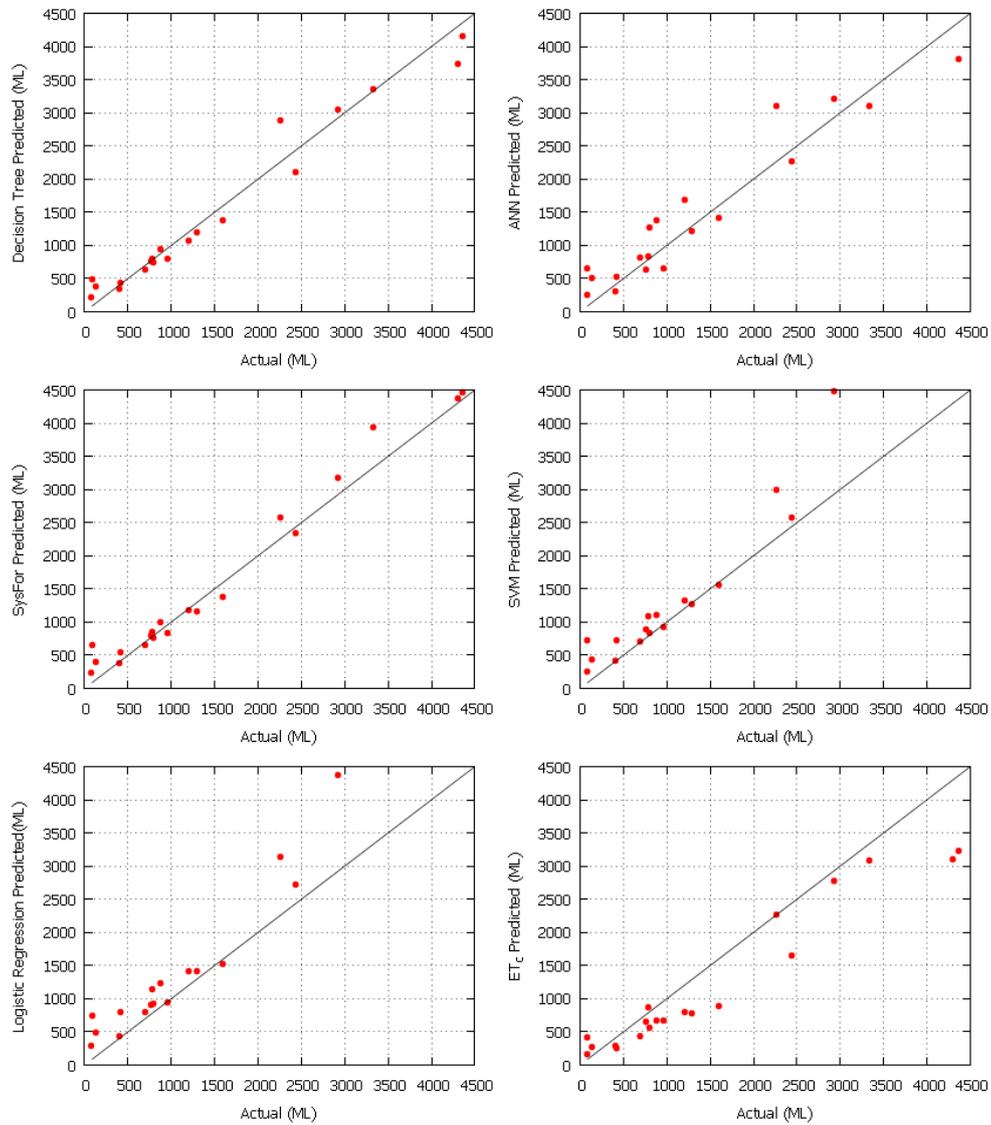

**Figure 7: Actual Vs Predicted Water Usage made by six different models on 22 nodes of CIA**

We also developed a web based Decision Support System (DSS) called Coleambally Integrated River Information System (Coleambally IRIS) which consists of a database and collection of various models (Khan et al, 2011). Users (farmers and irrigation managers) access various data from DSS including water predictions made by our model as shown in Figure 8. Based on our previous study we incorporated Decision Tree model in our DSS for predicting future water requirements. By using demand forecast results users will learn the water requirement for their particular farm for 7days in advance and can order the exact amount of water they require, this will increase the percentage of water savings and improve water use efficiency.

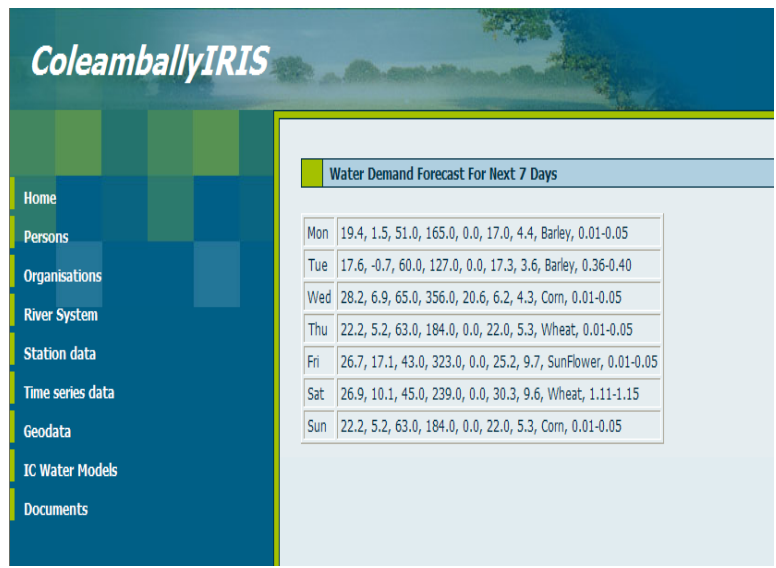

Figure 8: Irrigation water demand forecasting for 7 days

## 5. CONCLUSION

In this study we prepare a dataset, pre-process the dataset, apply and compare the effectiveness and performance of several data mining classification methods such as decision tree, ANN, SysFor, SVM and logistic regression for predicting irrigation water requirement. The novelty of this study is pre-processing the dataset using the combination of knowledge in both irrigation engineering and data mining, and comparison of SysFor with other classification techniques.

Our experimental results indicate a minor difference in the prediction accuracies achieved by different data mining techniques mainly SysFor, Decision tree and ANN. Computational results demonstrate that based on 3 folds cross validation method multiple decision tree technique SysFor produce the best prediction accuracy of 78% followed by decision tree and SVM with 74% and 64% respectively.

We also compare the prediction accuracies of the models with the actual water consumed by the crop. The closeness of prediction accuracy of SysFor performs the best with 97.5% followed by decision tree with 96% accuracy. Interestingly, ANN performs better than SVM by closely predicting the water demand to actual water used with 95% accuracy. The accuracy predictions made by SVM, logistic regression and traditional ETc method are found to be 78%, 75% and 77% respectively.

Therefore, from the above results we recommend that SysFor, decision tree and ANN techniques are most suitable for predicting irrigation water requirement. By developing and implementing a demand forecasting model using these techniques the farmers and irrigation managers of CIA can learn the future water requirement in advance accurately. Hence, this tool is crucial for effectively improving existing water management practices and maximising water productivity. Although the results obtained from this study are more significant for predicting water demand, the limitation would be use of less number of influential attributes in the dataset. This can be further improved by adding more attributes having high influence on crop water usage such as seepage, soil moisture, etc. In addition it would be interesting to explore the influence of cropping stage on crop water use. Furthermore, based on our results from this study we plan to incorporate SysFor model into our DSS to make the water predictions more accurate and reliable.